\RequirePackage[OT1]{fontenc}
\documentclass[a4paper, 10pt, conference]{ieeeconf}      

\IEEEoverridecommandlockouts                  

\usepackage[british]{babel}
\usepackage{amsmath,amssymb,amsfonts}
\usepackage{csquotes}
\usepackage{url}
\usepackage{graphicx}
\usepackage{amstext} 
\usepackage{array}   
\usepackage{makecell}
\usepackage{booktabs}

\RequirePackage[dvipsnames]{xcolor}       

\RequirePackage{tabularx}
\RequirePackage{booktabs}

\RequirePackage{graphicx} 
\RequirePackage{subcaption}   

\usepackage{tikz}
\usepackage{tikzscale}
\usepackage{tikzpeople}
\usepackage{tikz-3dplot}
\usepackage{pgfplots}
\usetikzlibrary{
  arrows,
  arrows.meta,
  backgrounds, 
  calc, 
  chains,
  decorations.pathreplacing,
  fit, 
  math,
  matrix,
  patterns, 
  positioning, 
  shadows.blur,
  shapes, 
  shapes.arrows,
  shapes.multipart
  }
\usepgfplotslibrary{groupplots,dateplot}
\pgfplotsset{compat=newest}
\usepgfplotslibrary{polar}

\definecolor{color_street}{RGB}{124,124,124}
\definecolor{color_border}{RGB}{56,56,56}
\definecolor{color_sidewalk}{RGB}{180,180,180}
\definecolor{light_blue}{RGB}{179, 217, 255} 
\definecolor{light_red}{RGB}{255, 153, 128}
\definecolor{light_green}{RGB}{173, 235, 173}

\definecolor{coco_joint_color_1}{RGB}{192,192,192}
\definecolor{coco_joint_color_2}{RGB}{255,255,0}
\definecolor{coco_joint_color_3}{RGB}{255,0,255}
\definecolor{coco_joint_color_4}{RGB}{255,255,102}
\definecolor{coco_joint_color_5}{RGB}{255,102,255}
\definecolor{coco_joint_color_6}{RGB}{0,255,0}
\definecolor{coco_joint_color_7}{RGB}{255,0,0}
\definecolor{coco_joint_color_8}{RGB}{51,255,51}
\definecolor{coco_joint_color_9}{RGB}{255,51,51}
\definecolor{coco_joint_color_10}{RGB}{102,255,102}
\definecolor{coco_joint_color_11}{RGB}{255,102,102}
\definecolor{coco_joint_color_12}{RGB}{0,102,0}
\definecolor{coco_joint_color_13}{RGB}{102,0,0}
\definecolor{coco_joint_color_14}{RGB}{0,153,0}
\definecolor{coco_joint_color_15}{RGB}{153,0,0}
\definecolor{coco_joint_color_16}{RGB}{0,204,0}
\definecolor{coco_joint_color_17}{RGB}{204,0,0}

\definecolor{coco_limb_color_1}{RGB}{252, 157, 154}
\definecolor{coco_limb_color_2}{RGB}{200, 255, 0}
\definecolor{coco_limb_color_3}{RGB}{252, 157, 154}
\definecolor{coco_limb_color_4}{RGB}{252, 157, 154}
\definecolor{coco_limb_color_5}{RGB}{200, 255, 0}
\definecolor{coco_limb_color_6}{RGB}{200, 255, 0}
\definecolor{coco_limb_color_7}{RGB}{124, 244, 154}
\definecolor{coco_limb_color_8}{RGB}{250, 2, 60}
\definecolor{coco_limb_color_9}{RGB}{250, 2, 60}
\definecolor{coco_limb_color_10}{RGB}{250, 2, 60}
\definecolor{coco_limb_color_11}{RGB}{124, 244, 154}
\definecolor{coco_limb_color_12}{RGB}{124, 244, 154}
\definecolor{coco_limb_color_13}{RGB}{252, 157, 154}
\definecolor{coco_limb_color_14}{RGB}{252, 157, 154}
\definecolor{coco_limb_color_15}{RGB}{200, 255, 0}
\definecolor{coco_limb_color_16}{RGB}{200, 255, 0}

\definecolor{pedrec_joint_color_1}{RGB}{192,192,192}  
\definecolor{pedrec_joint_color_2}{RGB}{255,255,0}    
\definecolor{pedrec_joint_color_3}{RGB}{255,0,255}    
\definecolor{pedrec_joint_color_4}{RGB}{255,255,102}  
\definecolor{pedrec_joint_color_5}{RGB}{255,102,255}  
\definecolor{pedrec_joint_color_6}{RGB}{0,255,0}      
\definecolor{pedrec_joint_color_7}{RGB}{255,0,0}      
\definecolor{pedrec_joint_color_8}{RGB}{51,255,51}    
\definecolor{pedrec_joint_color_9}{RGB}{255,51,51}    
\definecolor{pedrec_joint_color_10}{RGB}{102,255,102} 
\definecolor{pedrec_joint_color_11}{RGB}{255,102,102} 
\definecolor{pedrec_joint_color_12}{RGB}{0,102,0}     
\definecolor{pedrec_joint_color_13}{RGB}{102,0,0}     
\definecolor{pedrec_joint_color_14}{RGB}{0,153,0}     
\definecolor{pedrec_joint_color_15}{RGB}{153,0,0}     
\definecolor{pedrec_joint_color_16}{RGB}{0,204,0}     
\definecolor{pedrec_joint_color_17}{RGB}{204,0,0}     
\definecolor{pedrec_joint_color_18}{RGB}{0,0,0}       
\definecolor{pedrec_joint_color_19}{RGB}{51,51,51}    
\definecolor{pedrec_joint_color_20}{RGB}{101,101,101} 
\definecolor{pedrec_joint_color_21}{RGB}{148,148,148} 
\definecolor{pedrec_joint_color_22}{RGB}{224,224,224} 
\definecolor{pedrec_joint_color_23}{RGB}{0,255,0}     
\definecolor{pedrec_joint_color_24}{RGB}{255,0,0}     
\definecolor{pedrec_joint_color_25}{RGB}{152,255,152} 
\definecolor{pedrec_joint_color_26}{RGB}{255,152,152} 

\definecolor{pedrec_limb_color_1}{RGB}{252,157,154}  
\definecolor{pedrec_limb_color_2}{RGB}{252,157,154}  
\definecolor{pedrec_limb_color_3}{RGB}{200,255,0}    
\definecolor{pedrec_limb_color_4}{RGB}{200,255,0}    
\definecolor{pedrec_limb_color_5}{RGB}{250,2,60}     
\definecolor{pedrec_limb_color_6}{RGB}{250,2,60}     
\definecolor{pedrec_limb_color_7}{RGB}{124,244,154}  
\definecolor{pedrec_limb_color_8}{RGB}{124,244,154}  
\definecolor{pedrec_limb_color_9}{RGB}{252,157,154}  
\definecolor{pedrec_limb_color_10}{RGB}{252,157,154} 
\definecolor{pedrec_limb_color_11}{RGB}{200,255,0}   
\definecolor{pedrec_limb_color_12}{RGB}{200,255,0}   
\definecolor{pedrec_limb_color_13}{RGB}{124,244,154} 
\definecolor{pedrec_limb_color_14}{RGB}{250,2,60}    
\definecolor{pedrec_limb_color_15}{RGB}{124,124,124} 
\definecolor{pedrec_limb_color_16}{RGB}{124,124,124} 
\definecolor{pedrec_limb_color_17}{RGB}{124,124,124} 
\definecolor{pedrec_limb_color_18}{RGB}{124,124,124} 
\definecolor{pedrec_limb_color_19}{RGB}{200,255,0}   
\definecolor{pedrec_limb_color_20}{RGB}{252,157,154} 
\definecolor{pedrec_limb_color_21}{RGB}{200,200,200} 
\definecolor{pedrec_limb_color_22}{RGB}{124,244,154} 
\definecolor{pedrec_limb_color_23}{RGB}{250,2,60}    
\definecolor{pedrec_limb_color_24}{RGB}{200,255,0}   
\definecolor{pedrec_limb_color_25}{RGB}{252,157,154} 

\definecolor{coco_skel_1}{RGB}{255, 0, 0}
\definecolor{coco_skel_2}{RGB}{0, 255, 0}

\pgfkeys{tikz/matrixgrid/.style={
  matrix of nodes,
  nodes in empty cells,
  column sep      = -\pgflinewidth,
  row sep         = -\pgflinewidth,
  nodes={%
    inner sep=0mm,outer sep=0pt,
    minimum size=6mm,
    text height=\ht\strutbox,text depth=\dp\strutbox,
    draw
  }
}}

\pgfkeys{tikz/matrixbrace/.style={decorate, decoration={brace}, thick}}

\pgfkeys{tikz/matrix-area-marker/.style={
  fill,
  opacity=0.2,
  rounded corners,
  inner sep=-1pt,
  line width=1pt
}}

\pgfkeys{tikz/matrix-delimiter/.style={
  left delimiter=[,
  right delimiter=],
  inner sep=-2pt
}}

\pgfkeys{tikz/matrix-marker/.style={
  line cap= round,
  line width =15pt,
  opacity=0.2
}}

\pgfkeys{tikz/matrix-circle-marker/.style={
  fill,
  opacity=0.2,
  circle,
  inner sep=-1pt,
  line width=1pt
}}

\pgfdeclarelayer{bg}
\pgfdeclarelayer{layer1}
\pgfdeclarelayer{layer2}
\pgfdeclarelayer{layer3}
\pgfsetlayers{background,bg,main,layer1,layer2,layer3}

\newcommand\pgfmathsinandcos[3]{%
  \pgfmathsetmacro#1{sin(#3)}%
  \pgfmathsetmacro#2{cos(#3)}%
} 
\newcommand\LongitudePlane[3][current plane]{%
  \pgfmathsinandcos\sinEl\cosEl{#2} 
  \pgfmathsinandcos\sint\cost{#3} 
  \tikzset{#1/.estyle={cm={\cost,\sint*\sinEl,0,\cosEl,(0,0)}}}
}
\newcommand\LatitudePlane[3][current plane]{%
  \pgfmathsinandcos\sinEl\cosEl{#2} 
  \pgfmathsinandcos\sint\cost{#3} 
  \pgfmathsetmacro\yshift{\cosEl*\sint}
  \tikzset{#1/.estyle={cm={\cost,0,0,\cost*\sinEl,(0,\yshift)}}} %
}
\NewDocumentCommand{\DrawLongitudeCircle}{O{1} O{black} m}{
  \LongitudePlane{\angEl}{#3}
  \tikzset{current plane/.prefix style={scale=#1}}
  \pgfmathsetmacro\angVis{atan(sin(#3)*cos(\angEl)/sin(\angEl))} %
  \draw[current plane] (\angVis:1) arc (\angVis:\angVis+180:1); 
  \draw[current plane,dashed] (\angVis-180:1) arc (\angVis-180:\angVis:1); 
}
\NewDocumentCommand{\DrawLatitudeCircle}{ O{1} O{black} O{0.5} m }{
  \LatitudePlane{\angEl}{#4}
  \tikzset{current plane/.prefix style={scale=#1}}
  \pgfmathsetmacro\sinVis{sin(#4)/cos(#4)*sin(\angEl)/cos(\angEl)} 
  \pgfmathsetmacro\angVis{asin(min(1,max(\sinVis,-1)))}  
  \draw[current plane,#2, line width=#3] (\angVis:1) arc (\angVis:-\angVis-180:1); 
  \draw[current plane,dashed, #2, line width=#3] (180-\angVis:1) arc (180-\angVis:\angVis:1);
}

\makeatletter \newcommand{\pgfplotsdrawaxis}{\pgfplots@draw@axis} \makeatother
\pgfplotsset{axis line on top/.style={
  axis line style=transparent,
  ticklabel style=transparent,
  tick style=transparent,
  axis on top=false,
  after end axis/.append code={
    \pgfplotsset{axis line style=opaque,
      ticklabel style=opaque,
      tick style=opaque,
      grid=none}
    \pgfplotsdrawaxis}
  }
}

\makeatletter
\long\def\ifnodedefined#1#2#3{%
    \@ifundefined{pgf@sh@ns@#1}{#3}{#2}%
}
\makeatother

\newcolumntype{L}{>{$}l<{$}}
\DeclareGraphicsExtensions{.pdf,.png,.jpg,.jpeg}
\graphicspath{{figures/}{pictures/}{images/}{./}} 

\title{\LARGE \bf
PedRecNet: Multi-task deep neural network for full 3D human pose and orientation estimation
}

\author{Dennis Burgermeister and Crist\'obal Curio\\
	\scriptsize \\
\thanks{D. Burgermeister and C. Curio are with the Cognitive Systems Group, Computer Science Department, Reutlingen University, Germany.
{\tt\small \{Dennis.Burgermeister, Cristobal.Curio\}@Reutlingen-University.de}
}}

\begin{document}

\maketitle

\thispagestyle{empty}
\pagestyle{empty}

\begin{abstract}
We present a multitask network that supports various deep neural network based pedestrian detection functions. Besides 2D and 3D human pose, it also supports body and head orientation estimation based on full body bounding box input. This eliminates the need for explicit face recognition. We show that the performance of 3D human pose estimation and orientation estimation is comparable to the state-of-the-art. Since very few data sets exist for 3D human pose and in particular body and head orientation estimation based on full body data, we further show the benefit of particular simulation data to train the network. The network architecture is relatively simple, yet powerful, and easily adaptable for further research and applications.
\end{abstract}

\section{Introduction}
The detection of pedestrians as well as their behavior remains a challenge in the field of autonomous driving. Previous work shows how to detect pedestrian actions using 2D pose recognition~\cite{ludlSimpleEfficientRealtime2019}. In this work, we present a new multitask network, PedRecNet, that can estimate 3D poses in addition to 2D poses. 3D poses bring in the benefit of multiple perspectives on the skeleton, enabling detection of movement changes which may not be visible in 2D projections~\cite{bulthoffTopdownInfluencesStereoscopic1998}. In addition to 3D pose data, orientation information about a person's body and head is also relevant for human recognition systems. Especially in pedestrian recognition, this information can be valuable to perform path planning or to detect if a pedestrian notices a vehicle or not. Since 3D pose recognition and body as well as head orientation estimation are related it seems beneficial to bring this tasks together in one versatile network. All tasks in the PedRecNet are based on the same input data, so all tasks should be implemented in the same network using a multitask approach. Since there are only a few real datasets available for 3D pose recognition and especially for body and head orientation estimation, we use simulation data to improve these parts of the PedRecNet and to enable training in the first place. The skeleton-based action recognition results presented by \cite{ludlSimpleEfficientRealtime2019} and \cite{ludlEnhancingDataDrivenAlgorithms2020} support the assumption that using abstract pose information rather than just visual information enables the transfer of simulated training data to real data. We will corroborate this work hypothesis in more detail in experiments using simulated training data. In the following, we describe the developed network, the datasets used, and evaluate 3D human pose recognition as well as the body orientation estimation on several real and simulated datasets.

The entire system and novel simulation data has been made public under the MIT license\footnote{\url{https://github.com/noboevbo/PedRec} (accessed on 2021-09-02)}.

Our contributions in this work are:

\begin{enumerate}
\item A multi-task network which supports 2D and 3D human pose estimation, as well as body and head orientation estimation on cropped full body input data.
\item An approach to 3D human pose estimation in which 3D human joint positions are encoded in the skeletal coordinate system. This makes the skeleton estimation independent of the camera parameters and can thus be better used in follow-up applications, such as action recognition which uses temporal data.
\item An integrated approach to body and head orientation estimation based on whole body bounding box input. This eliminates the need for face recognition to obtain a crop of the head bounding box.
\item Simulation data that provide, in particular, accurate head and body orientation data that are not available in standard data sets.
\end{enumerate}

\section{Related work}
Besides direct estimation of 3D human joint positions by a deep neural network~\cite{mehtaVNectRealtime3D2017,mehtaSingleShotMultiperson3D2018,mehtaXNectRealtimeMultiperson2020,luvizonMultitaskDeepLearning2020} alternative approaches exist that first estimate 2D poses followed by 3D pose regression~\cite{chen3DHumanPose2017,pavllo3DHumanPose2019,tekinLearningFuse2D2017,zhou3DHumanPose2017,kolotourosLearningReconstruct3D2019}. Some approaches show the application of model-based approaches~\cite{bogoKeepItSMPL2016,zhouDeepKinematicPose2016,mehtaVNectRealtime3D2017,mehtaXNectRealtimeMultiperson2020,kolotourosLearningReconstruct3D2019} to further improve a recognized 3D skeleton. The categorization in bottom-up~\cite{mehtaSingleShotMultiperson3D2018} and top-down approaches~\cite{luvizon2D3DPose2018,mehtaVNectRealtime3D2017,mehtaXNectRealtimeMultiperson2020} is also valid for 3D human pose estimation. Mehta~\textit{et al.} show an approach predicting three location-maps for the $x$, $y$ and $z$ position parameters per body joint \cite{mehtaVNectRealtime3D2017}. Those location maps encode the distance in $x$, $y$, or $z$ direction from the coordinate root (hip center). The location of a joint on those location maps is retrieved from 2D pose heatmaps. The results are refined using a kinematic skeleton fitting method. They have also shown how to apply location maps in a bottom-up approach which also handles occlusion better by using redundancy in so-called occlusion-robust pose-maps by representing the decomposed body as torso, limbs, and heads \cite{mehtaSingleShotMultiperson3D2018}. Luvizon \textit{et al.} show a similar approach to encode depth information in a heatmap but use it only for the depth estimation~\cite{luvizonMultitaskDeepLearning2020}. It is also possible to directly regress 3D poses from 2D heatmaps \cite{tekinLearningFuse2D2017} or directly from 2D pose coordinates \cite{martinezSimpleEffectiveBaseline2017} which improves when using multiple frames as input \cite{pavllo3DHumanPose2019}. Another approach to retrieve 3D pose information from 2D poses is 3D catalog matching \cite{chen3DHumanPose2017}. Such approaches rely more heavily on the 2D human pose estimator's output than approaches that also use visual input. Kolotouros \textit{et al.} show how to reconstruct a volumetric model by estimating the parameters for the SMPL statistical body shape model~\cite{loperSMPLSkinnedMultiperson2015} and further improve the model by iteratively fitting on 2D joints~\cite{kolotourosLearningReconstruct3D2019}. We present an approach similar to \cite{luvizonMultitaskDeepLearning2020}, but with a straightforward and performant method to retrieve the pose and depth heatmaps in section \ref{sec:architecture}.

Body and head orientation estimation approaches are usually handled as separate problems. The estimation of body orientation often originate in pedestrian-related works. Classical approaches often used classification of body parts, for example, by using a part descriptor in a sliding window fashion to classify position, scale, and orientation of body parts \cite{andrilukaPictorialStructuresRevisited2009}. Another approach focuses on combining pedestrian orientation and classification by clustering pedestrians in the four categories front, back, left, and right and train classification networks on those clusters, which combined scores serve as the full pedestrian classification \cite{enzweilerIntegratedPedestrianClassification2010}. Another approach uses specific detectors for head and body orientation which are converted to a full continuous probability density function and stabilized over time by particle filtering \cite{flohrJointProbabilisticPedestrian2014,flohrProbabilisticFrameworkJoint2015}. The authors also discretized the orientation space to 45-degree bins and used a HOG/linSVM based classification system \cite{flohrJointProbabilisticPedestrian2014}. There is a lot of recent head orientation work using deep learning \cite{guptaNoseEyesEars2019,ruizFineGrainedHeadPose2018,panSelfPacedDeepRegression2020,huDeepConvolutionalNeural2021,valleMultitaskHeadPose2020,xiaEfficientMultitaskNeural2021}, which usually takes a head bounding box as input and thus is an additional step in the recognition pipeline. Such methods require a head bounding box with a reasonable resolution and thus high-resolution sensors or people not at distance from the camera sensor. Work on body orientation or combined body and head orientation estimation has not yet transitioned well to deep learning approaches. One possible reason could be the lack of appropriate datasets. There are not many standard datasets, and the existing ones are rather small and thus not suitable to train deep neural networks. Heo~\textit{et al.} try to overcome this issue on body orientation estimation by using a teacher-student learning framework in which they train a teacher network with labeled data and use this network to generate labels for an unlabeled dataset with which the student network is trained \cite{heoEstimationPedestrianPose2019}. They have also discretized the output orientation in 45-degree bins, turning the problem into a classification problem \cite{heoEstimationPedestrianPose2019}. Another work uses CNNs in a random forest that focuses on different body and head parts to recognize the human body and head orientation, with a focus on head orientation \cite{leeHeadBodyOrientation2019}. Steinhoff and Göhring propose the usage of IMUs to generate more labeled training data for body and head orientation tasks, but IMU-based approaches are usually hard to sync, suffer from error accumulation, and do not contain global reference points \cite{steinhoffPedestrianHeadBody2020}. Wu~\textit{et al.} propose the application of 3D human pose estimation approaches as a basis for body and head pose estimation using a 3D pose estimation network as a backbone for a classification header which classifies the input in $72$ orientation bins \cite{wuMEBOWMonocularEstimation2020}. We propose a regression approach which is based on full human 3D pose information, described in section \ref{sec:architecture}. Regressed orientation estimation offer various benefits, for example in time-dependent fine-grained actions like the head movement during looking for traffic. We show how to train such a network with simulated data to overcome the deficient number of labeled data in this field. 

\section{Method}
\label{sec:architecture}
\begin{figure}[bth] 
  \centering    
  \resizebox{1\columnwidth}{!}{\input{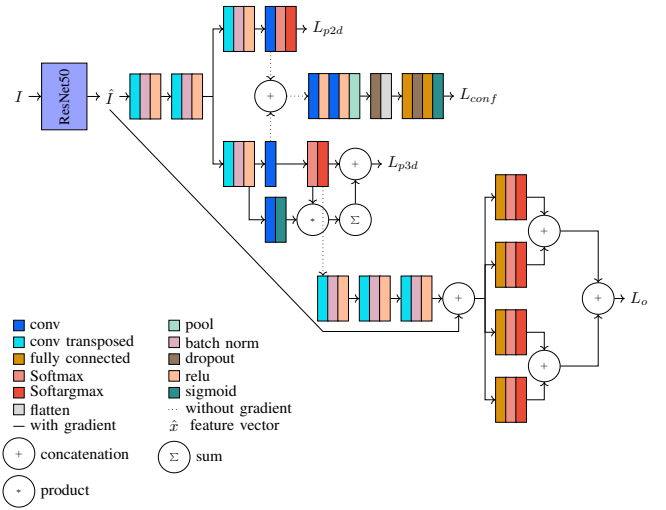}}
  \caption[Simplified PedRecNet architecture]{Simplified PedRecNet architecture. The input $I$ is an RGB cropped bounding box image of a human. The dotted connection lines indicate connections without gradient flow in the backward pass.}
  \label{fig:pedrecnet}
\end{figure}

The overall network architecture is shown in Figure \ref{fig:pedrecnet}. The PedRecNet expands the 2D human pose estimation approach proposed by Xiao~\textit{et al.}~\cite{xiaoSimpleBaselinesHuman2018}. The PedRecNet architecture is based on a ResNet50 backbone for feature extraction but other backbones could be used as well. The ResNet50 architecture was chosen as a compromise between accuracy and performance. The inputs $I$ are always images cropped to the size of certain bounding boxes. First features $\hat{I}$ are extracted using the feature extraction part of the network. Next, the 2D human pose estimation part is based on three transpose convolution blocks with which joint heatmaps are generated from the extracted features~\cite{xiaoSimpleBaselinesHuman2018}. 

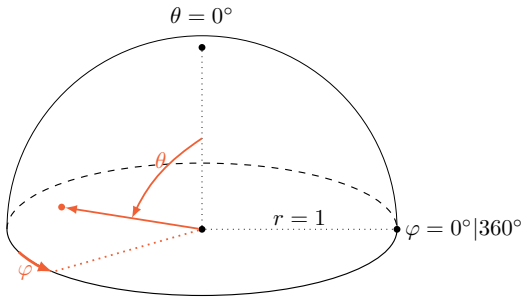
\begin{figure}[bth] 
  \centering    
  \resizebox{0.8\columnwidth}{!}{\begin{tikzpicture} 
  \tikzset{%
  >=latex, 
  inner sep=0pt,%
  outer sep=2pt,%
  mark coordinate/.style={inner sep=0pt,outer sep=0pt,minimum size=3pt,fill=black,circle}%
}

\def\radius{3} 
\def\angEl{20} 
\def\angAz{0} 
\def\angPhiOne{220} 
\def\angThetaOne{20} 


\pgfmathsetmacro\H{\radius*cos(\angEl)} 
\LongitudePlane[xzplane]{\angEl}{\angAz}
\LongitudePlane[pzplane]{\angEl}{\angPhiOne}
\LatitudePlane[equator]{\angEl}{0}


\draw (-3,0) arc(180:0:3);

\DrawLatitudeCircle[\radius]{0} 

\coordinate[mark coordinate] (origin) at (0,0);
\path[xzplane] (\radius,0) coordinate[mark coordinate] (PHI_0) node[right=0 of PHI_0] () {$\varphi = 0^\circ | 360^\circ$};
\path[xzplane] (0,\radius) coordinate[mark coordinate] (THETA_0) node[above=0.25 of THETA_0] () {$\theta = 0^\circ$};
\draw[dotted] (origin) -- (PHI_0) node[midway, above] (MID_PHI_0) {$r = 1$};
\draw[dotted] (origin) -- (THETA_0) coordinate[midway] (MID_THETA_0) coordinate[pos=0.55] (NEAR_START_THETA_0);

\path[pzplane] (\angThetaOne:\radius) coordinate[mark coordinate, RedOrange] (C1);
\path[pzplane] (\radius, 0) coordinate (C1_equator);
\draw[dotted, RedOrange, thick] (origin) -- (C1_equator);
\draw[->, RedOrange, thick] (origin) -- (C1) coordinate[midway] (MID_C1);
\draw[->,thin, RedOrange, thick] (MID_THETA_0) to[bend right=15] node[midway,above] {$\theta$} (MID_C1);
\draw[equator,->, RedOrange, very thick] (\angPhiOne - 20:\radius) arc (\angPhiOne - 20:\angPhiOne:\radius) node[left=1.5ex] {$\varphi$};



\end{tikzpicture} }
  \caption{Visualization of the orientation estimation for each, the body and the head. The orange dot shows an example point on a 3D sphere, visualizing an orientation. We use the standard notation from ISO 80000-2:2019\cite{internationalorganizationforstandardizationISO80000220192019} for spherical coordinates. As we only need the polar angle $\theta$ and azimuthal angle $\varphi$ we use a unit sphere and with $r=1$.}
  \label{fig:camera_sphere}
\end{figure}

In PedRecNet, the 2D human pose estimation architecture was extended to include 3D human pose estimation. For this purpose, two transpose convolution blocks are used as a common basis and then split into a 2D and a 3D path. These have basically the same structure. The output in the 2D path corresponds to 2D image coordinates. The 3D path, leading to $L_{p3d}$, corresponds to the estimation of the $x$ and $y$ coordinates of a joint relative to the hip and additional depth estimation of the $z$ coordinate using a sigmoid map. Another change from the previous 2D pose estimation approach is the post-processing of the heatmaps. In the approach shown in \cite{ludlSimpleEfficientRealtime2019}, the heatmaps were output from the network, and a non-maximum suppression (NMS) was used to determine the coordinates. This has the disadvantage that the subpixel coordinate values are not available in the network. The NMS approach also implies that artificial heatmaps have to be generated to be used as training data labels. The PedRecNet applies a softargmax layer, which determines the coordinates from the heatmaps in the network inside the network. This allows to use estimated joint coordinates as inputs into other parts of the network. For example, the 3D joint positions are used as input into the orientation estimation part of the network, which is visualized as the path to $L_{o}$. The 3D joint position coordinates are scaled up into a feature space via 1D transpose convolution blocks. These are concatenated with the visual features $\hat{I}$ such that the orientation estimation can use direct information from the image in addition to the, noisy and potentially erroneous, 3D pose. This concatenated feature vector is input into a fully connected layer which generates a one-dimensional heatmap for each, the polar angle $\theta$ and azimuthal angle $\varphi$ (see Figure~\ref{fig:camera_sphere}) of the body as well as the head. From these one-dimensional orientation heatmaps, the corresponding normalized angle is extracted using 1D softargmax. To classify the visibility of labels, we added a standard classification head to the network leading to the path to $L_{conf}$.

The PedRecNet outputs 2D human joint positions as pixel coordinates in the bounding box's coordinate system, normalized between zero and one. The 3D human joint positions are output as 3D coordinates relative to the hip center position, and normalized between zero and one. Orientations are output as angles between $0-180^{\circ}$ for $\theta$ and $0-360^{\circ}$ for $\varphi$, and also normalized between zero and one.

\subsection{Datasets}
\label{sec:pedrec:datasets}
The training and validation datasets of the PedRec network are composed of the COCO~\cite{linMicrosoftCOCOCommon2014}, the H36M~\cite{ionescuHuman36MLarge2014}, and self-generated simulated datasets. The datasets support different labels, especially orientation estimation labels are only available in the simulation data (cf. Table~\ref{tab:pedrec_dataset_labels}).

\begin{table}[!htbp]
  \centering
  \begin{tabular}{l l l l} \toprule
      Dataset & Pose2D & Pose3D & Orientation \\ \midrule
      COCO & \checkmark & \text{\sffamily X} & \text{\sffamily O} \\
      H36M & \checkmark & \checkmark & \text{\sffamily X} \\
      TUD~\cite{andrilukaMonocular3DPose2010} & \text{\sffamily X} & \text{\sffamily X} & \text{\sffamily O}\\
      SIM-ROM & \checkmark & \checkmark & \checkmark \\
      SIM-Circle & \checkmark & \checkmark & \checkmark \\
  \end{tabular}
  \caption[Dataset Label Support]{Overview of used datasets and the supported labels. COCO (MEBOW~\cite{wuMEBOWMonocularEstimation2020}) and TUD~\cite{andrilukaMonocular3DPose2010} orientation annotations provide only body $\varphi$ labels. The 2D and 3D pose estimation labels differ in the available labels as COCO, H36M and SIM use different skeleton structures. Legend: \checkmark data is available, \text{\sffamily X} data is not available, {\sffamily O} data is partially available.}
  \label{tab:pedrec_dataset_labels}
\end{table}

Whole-body data, including the $\theta$ and $\varphi$ angle of the body and head orientations, are only available in the simulated datasets. Therefore, the validation of the orientation estimation on real data can be performed with these datasets only for the azimuthal angle $\varphi$ of the body. All simulated datasets are created using motions, captured in a motion capture laboratory.

\subsubsection{SIM-ROM}
In the SIM-ROM dataset, a person was recorded performing an extended range of motions that should include as many poses as possible. The idea behind this dataset is to provide data from various performable body poses for 3D pose recognition.

\subsubsection{SIM-C01}
The SIM-C01 dataset is a large scale pedestrian action dataset containing actions ranging from simple walking, to hitchhiking, to tripping and falling. This dataset is used in this work for validation (SIM-C01V) only.

\subsubsection{SIM-CIRCLE}
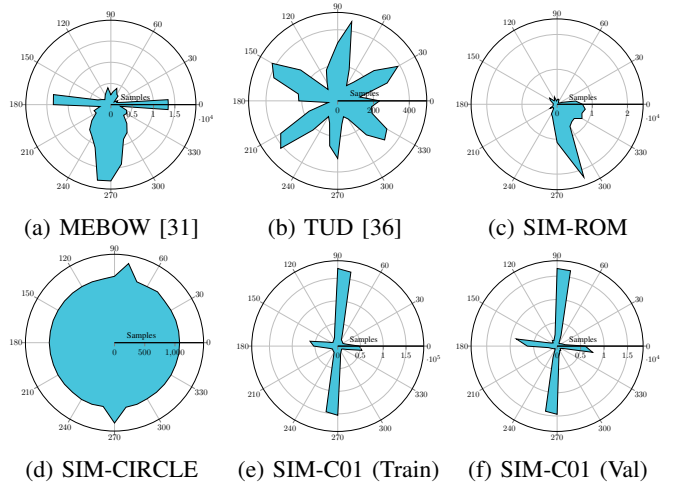
\begin{figure}[!htbp]
  \centering
  \subcaptionbox{MEBOW~\cite{wuMEBOWMonocularEstimation2020}}[.32\columnwidth]{%
  \resizebox{0.32\columnwidth}{!}{%
  \begin{tikzpicture}
  \definecolor{color0}{rgb}{0.12156862745098,0.466666666666667,0.705882352941177}
  
  \begin{polaraxis}[
  ylabel={Samples},
  ylabel style={yshift=-5pt,anchor=east},
  axis line on top,
  yticklabel style={color=black,
  ytick align=outside,
  anchor=north,
        yshift=-2*\pgfkeysvalueof{/pgfplots/major tick length}}
  ]
  \addplot [no markers, very thin, fill=SkyBlue] table [col sep=comma] {code/coco_mebow_body_orientations.csv};
  \end{polaraxis}

\end{tikzpicture}}
  }\hfill
  \subcaptionbox{TUD~\cite{andrilukaMonocular3DPose2010}}[.32\columnwidth]{%
  \resizebox{0.32\columnwidth}{!}{%
    \begin{tikzpicture}

  \definecolor{color0}{rgb}{0.12156862745098,0.466666666666667,0.705882352941177}
  
  \begin{polaraxis}[
  width=0.5\textwidth,
  ylabel={Samples},
  ylabel style={yshift=-5pt,anchor=east},
  axis line on top,
  yticklabel style={color=black,
  ytick align=outside,
  anchor=north,
        yshift=-2*\pgfkeysvalueof{/pgfplots/major tick length}}
  ]
  \addplot [no markers, very thin, fill=SkyBlue] table [col sep=comma] {code/tud_body_orientations.csv};
  \end{polaraxis}
  
\end{tikzpicture}}
  }\hfill
  \subcaptionbox{SIM-ROM}[.32\columnwidth]{%
  \resizebox{0.32\columnwidth}{!}{%
    \begin{tikzpicture}

  \definecolor{color0}{rgb}{0.12156862745098,0.466666666666667,0.705882352941177}
  
  \begin{polaraxis}[
  width=0.5\textwidth,
  ylabel={Samples},
  ylabel style={yshift=-5pt,anchor=east},
  axis line on top,
  yticklabel style={color=black,
  ytick align=outside,
  anchor=north,
        yshift=-2*\pgfkeysvalueof{/pgfplots/major tick length}}
  ]
  \addplot [no markers, very thin, fill=SkyBlue] table [col sep=comma] {code/rtsim_rom_body_orientations.csv};
  \end{polaraxis}
  
\end{tikzpicture}}
  } 
  \subcaptionbox{SIM-CIRCLE\label{fig:datasets_body_orientations:circle}}[.32\columnwidth]{%
  \resizebox{0.32\columnwidth}{!}{%
  \begin{tikzpicture}

  \definecolor{color0}{rgb}{0.12156862745098,0.466666666666667,0.705882352941177}
  
  \begin{polaraxis}[
  width=0.5\textwidth,
  ylabel={Samples},
  ylabel style={yshift=-5pt,anchor=east},
  axis line on top,
  yticklabel style={color=black,
  ytick align=outside,
  anchor=north,
        yshift=-2*\pgfkeysvalueof{/pgfplots/major tick length}}
  ]
  \addplot [no markers, very thin, fill=SkyBlue] table [col sep=comma] {code/circle01_body_orientations.csv};
  \end{polaraxis}
  
\end{tikzpicture}}
  }\hfill
  \subcaptionbox{SIM-C01 (Train)}[.32\columnwidth]{%
  \resizebox{0.32\columnwidth}{!}{%
    \begin{tikzpicture}

  \definecolor{color0}{rgb}{0.12156862745098,0.466666666666667,0.705882352941177}
  
  \begin{polaraxis}[
  width=0.5\textwidth,
  ylabel={Samples},
  ylabel style={yshift=-5pt,anchor=east},
  axis line on top,
  yticklabel style={color=black,
  ytick align=outside,
  anchor=north,
        yshift=-2*\pgfkeysvalueof{/pgfplots/major tick length}}
  ]
  \addplot [no markers, very thin, fill=SkyBlue] table [col sep=comma] {code/conti01_body_orientations.csv};
  \end{polaraxis}
  
\end{tikzpicture}}
  }
  \subcaptionbox{SIM-C01 (Val)}[.32\columnwidth]{%
  \resizebox{0.32\columnwidth}{!}{%
    \begin{tikzpicture}

  \definecolor{color0}{rgb}{0.12156862745098,0.466666666666667,0.705882352941177}
  
  \begin{polaraxis}[
  width=0.5\textwidth,
  ylabel={Samples},
  ylabel style={yshift=-5pt,anchor=east},
  axis line on top,
  yticklabel style={color=black,
  ytick align=outside,
  anchor=north,
        yshift=-2*\pgfkeysvalueof{/pgfplots/major tick length}}
  ]
  \addplot [no markers, very thin, fill=SkyBlue] table [col sep=comma] {code/conti01_val_body_orientations.csv};
  \end{polaraxis}
  
\end{tikzpicture}}
  }
  \caption[Distribution of body $\theta$ orientations]{Distribution of body $\theta$ orientations [$^\circ$] in the datasets used in this work. The plots show the distribution of the samples in a polar plot. The small peaks in SIM-CIRCLE are due to overlapping start and end frames.}
  \label{fig:datasets_body_orientations}
\end{figure}

The SIM-Circle dataset resulted from an analysis of the other datasets. As highlighted in Figure~\ref{fig:datasets_body_orientations}, there is a substantial difference in the distribution of the azimuthal angle $\varphi$ of the body pose.

The same could be observed for the distribution of the azimuthal angle $\varphi$ of the head poses. In order to generate further data with a uniform distribution, the SIM-Circle dataset was created, in which 3D models walk clock- and counterclockwise in a circle at a uniform speed (see Figure~\ref{fig:datasets_body_orientations:circle}). Table~\ref{tab:dataset_statistics} provides an overview over the datasets used.

\begin{table}[!htbp]
  \centering
  \resizebox{\columnwidth}{!}{
  \begin{tabular}{l l l l l l} \toprule
    Data & COCO~\cite{linMicrosoftCOCOCommon2014} & H36M~\cite{ionescuHuman36MLarge2014} & TUD~\cite{andrilukaMonocular3DPose2010} & SIM-ROM & SIM-CIRCLE \\ 
    \midrule
    $n$ & $149.813$ & $1.559.752$ & $8.322$ & $147.729$ & $39.484$ \\
    $\bar{b}^{2D} \; [px]$ & $229$ & $855$ & $193$ & $503$ & $222$ \\
    $\sigma_{b}^{2D} \; [px]$ & $148$ & $83$ & $44$ & $502$ & $133$ \\
    $\min b^{2D} \; [px]$ & $28$ & $585$ & $45$ & $31$ & $43$ \\
    $\max b^{2D} \; [px]$ & $1.002$ & $1.339$ & $462$ & $2.202$ & $830$ \\
    $\bar{b}^{3D} \; [mm]$ &  & $1.712$ &  & $1.816$ & $1.885$ \\
    $\sigma_{b}^{3D} \; [mm]$ &  & $197$ &  & $401$ & $194$\\
    $\min b^{3D} \; [mm]$ &  & $859$ & & $169$ & $1.202$\\
    $\max b^{3D} \; [mm]$ &  & $2.769$ & & $2.840$ & $2.194$ \\
    $\bar{d} \; [px]$ &  & $5.170$ & & $7.205$ & $10.609$ \\
    $\sigma_{d} \; [px]$ &  & $750$ & & $5.578$ & $5.779$\\
    $\min d \; [px]$ &  & $2.530$ & & $22$ & $2.321$ \\
    $\max d \; [px]$ &  & $7.690$ & & $21.289$ & $25.754$\\
  \end{tabular}}
  \caption[Dataset - Statistics]{Statistics of datasets used in the PedRec experiments. $b$ represents the 2D or 3D bounding box size and $d$ the distance to the camera. The number of samples is notated as $n$, the mean of bounding boxes and the distances to the camera as $\bar{b}/\bar{d}$, and the standard deviation values are notated with $\sigma$.}
  \label{tab:dataset_statistics}
\end{table}

This overview shows that the H36M dataset was recorded in a lab with limited space, which is why the bounding boxes are always relatively large. Comparing the 2D bounding box diameters of COCO, TUD, and SIM-CIRCLE, the distribution is similar, as each of the datasets contains individuals at various distances.

\subsection{Training procedure}
\label{sec:pedrec:training}
The network was trained step by step in the following order:

\begin{enumerate}
  \item 2D human pose estimation
  \item 3D human pose estimation
  \item Joint visibility
  \item Head and body orientation
\end{enumerate}

The training was performed on real data (COCO+H36M) and then on simulation data in the same order. 

\paragraph{Loss Functions}
\label{sec:pedrec:loss_functions}
We used the $L_1$ loss function for the 2D and 3D human joint coordinate regression losses $L_{p2d}$ and $L_{p3d}$. The joint visibility loss $L_{conf}$ is the standard binary cross-entropy loss. In the orientation regression task, we represent circular data in a one-dimensional map. Thus we cannot use the standard $L_1$ or $L_2$ loss; for example, a prediction of $359^{\circ}$ with a ground truth of $0^{\circ}$ results in an error of $359^{\circ}$. This applies only for the azimuthal angle $\varphi$, the polar angle $\theta$ is defined between $0^{\circ}$ and $180^{\circ}$, and thus the standard distance metrics can be used. As such, we applied the following loss functions:

\begin{align}
  L_\varphi &= \frac{\sum_{i=1}^{N} \min{(1 - |\hat{m_{\varphi}} - m_{\varphi}|, |\hat{m_{\varphi}} - m_{\varphi}|)}}{N} \label{eq:loss_o_phi}\\
  L_\theta &= \frac{\sum_{i=1}^{N} |\hat{m_{\theta}} - m_{\theta}|}{N} \label{eq:loss_o_theta}\\
  L_{o} &= \frac{L_\varphi + L_\theta}{2} \label{eq:loss_o}
\end{align}

where $N$ is the number of samples and $m_{\varphi}$ and $m_{\theta}$ are the softargmax outputs for the azimuthal angle $\varphi$ and the polar angle $\theta$ normalized between zero and one. As shown in equation~\ref{eq:loss_o_phi} and \ref{eq:loss_o_theta} the $L_1$ loss is applied. For training data which only provides labels for the azimuthal angle $\varphi$ only equation~\ref{eq:loss_o_phi} is used. 

In our experiments, $N$ may be a subset of the entire training set, as various datasets with different supported labels (see table~\ref{tab:pedrec_dataset_labels}) were combined. As such, each loss function contains a sample selection step before the actual calculation of the loss.

To weight the loss functions, we applied uncertainty loss, described by Kendall~\textit{et al.}~\cite{kendallMultitaskLearningUsing2018}, to balance the different loss outputs. The final loss function is:

\begin{equation}
  \resizebox{0.9\columnwidth}{!}{%
    $L = \frac{1}{2\sigma_1^2} L_{p2d} + \frac{1}{2\sigma_2^2} L_{p3d} + \frac{1}{2\sigma_3^2} L_{o} + \frac{1}{\sigma_4^2} L_{conf} + \log{(1+ \prod\limits_{i=1}^{4}\sigma_i)}$%
    }
\end{equation}

where $\sigma_{1-4}$ are learnable parameters. We use $\log{(1+\sigma)}$ instead $\log{(\sigma)}$ to ensure a positive loss value.

\paragraph{Optimizer}
For optimization we used the AdamW optimizer~\cite{loshchilovDecoupledWeightDecay2019} which is an slightly modified variant of the Adam optimizer~\cite{kingmaAdamMethodStochastic2015}. We applied the learning rate range test~\cite{smithCyclicalLearningRates2017a} to get an initial learning rate of $4e^{-3}$. We used a standard weight decay of $1e^{-2}$. We also applied the 1cycle policy~\cite{smithDisciplinedApproachNeural2018} with which the learning rate is updated during the training process from a minimum learning rate of $\frac{4e^{-3}}{25}$ to the maximum of $4e^{-3}$ and afterward back to a minimum using cosine annealing. Smith showed that this approach results in faster convergence and usually better results~\cite{smithCyclicalLearningRates2017a}. The network was trained with a training cycle of $15$ epochs, from which $10$ epochs were trained with frozen weights in the feature extractor. For the last five epochs with the feature extractor unfrozen, we reduce the learning rate for the feature extraction to $2e^{-4}$ and for the other layers to $4e^{-4}$. The training cycle was repeated five times, which improved the performance slightly.

\paragraph{Datasets}
We used the training datasets of COCO~\cite{linMicrosoftCOCOCommon2014}, H36M~\cite{ionescuHuman36MLarge2014}, SIM-ROM, and SIM-Circle to train the PedRecNet. The validation is done on the validation parts of COCO, H36M, and the SIM-C01 dataset. We subsampled the H36M training dataset by ten, all samples from the other datasets were used. For the orientation estimation part, we used only labels from SIM-ROM and SIM-Circle during the initial training. COCO labels were used in an additional training step during orientation experiments (see section \ref{sec:pedrec:results:orientation}). The dataset names are abbreviated in the results as follows: $C$ stands for COCO, $M$ for COCO (MEBOW~\cite{wuMEBOWMonocularEstimation2020}), $H$ for H36M and $S$ for SIM-Circle and SIM-ROM combined. COCO is always the base dataset of PedRecNet and is therefore used as one of the training dataset in every experiment.

\paragraph{Augmentations} We augmented the data by scaling the input by up to $\pm 25\%$. We rotate the input by up to $\pm 30^{\circ}$ in $50\%$ of the cases, but only when no orientation labels were used. The image is flipped in $50\%$ of the cases.

\section{Experiments}
\paragraph{3D pose estimation}
\begin{table*}[!htp]
  \centering
  \resizebox{\textwidth}{!}{%
  \begin{tabular}{l l l l l l l l l l l l l l l l l l}
    \toprule
    Method & Prop. & Dir. &  Disc. & Eat & Greet & Phone & Photo &  Pose & Purch. & Sit & SitD. & Smoke & Wait & WalkD. & Walk & WalkT. & Avg \\\midrule
    Chen~\textit{et al.} '17~\cite{chen3DHumanPose2017} & g & - & - & - & - & - & - & - & - & - & - & - & - & - & - & - & $82.4$ \\
    Martinez~\textit{et al.} '17~\cite{martinezSimpleEffectiveBaseline2017} & g & $51.8$ & $56.2$ & $58.1$ & $59.0$ & $69.5$ & $78.4$ & $55.2$ & $58.1$ & $74.0$ & $94.6$ & $62.3$ & $59.1$ & $65.1$ & $49.5$ & $52.4$ & $62.9$ \\
    Luvizon ~\textit{et al.} '17~\cite{luvizonHumanPoseRegression2017} & gs & $49.2$ & $51.6$ & $47.6$ & $50.5$ & $51.8$ & $48.9$ & $48.5$ & $51.7$ & $61.5$ & $70.9$ & $53.7$ & $60.3$ & $44.4$ & $48.9$ & $57.9$ & $53.2$ \\
    Yang~\textit{et al.} '18~\cite{yang3DHumanPose2018a} & - & $51.5$ & $58.9$ & $50.4$ & $57.0$ & $62.1$ & $65.4$ & $49.8$ & $52.7$ & $69.2$ & $85.2$ & $57.4$ & $58.4$ & $43.6$ & $60.1$ & $47.7$ & $58.6$ \\
    Pavllo~\textit{et al.} '18~\cite{pavllo3DHumanPose2019} & agt & $45.1$ & $47.4$ & $42.0$ & $46.0$ & $49.1$ & $56.7$ & $44.5$ & $44.4$ & $57.2$ & $66.1$ & $47.5$ & $44.8$ & $49.2$ & $32.6$ & $34.0$ & $47.1$ \\
    Pavllo~\textit{et al.} '18~\cite{pavllo3DHumanPose2019} & ag & $47.1$ & $50.6$ & $49.0$ & $51.8$ & $53.6$ & $61.4$ & $49.4$ & $47.4$ & $59.3$ & $67.4$ & $52.4$ & $49.5$ & $55.3$ & $39.5$ & $42.7$ & $51.8$ \\
    Luvizon~\textit{et al.} '20~\cite{luvizonMultitaskDeepLearning2020} & gs & $43.2$ & $48.6$ & $44.1$ & $45.9$ & $48.2$ & $43.5$ & $44.2$ & $45.5$ & $57.1$ & $64.2$ & $50.6$ & $53.8$ & $40.0$ & $44.0$ & $51.1$ & $\mathbf{48.6}$ \\ 
    Shan~\textit{et al.} '21~\cite{shanImprovingRobustnessAccuracy2021} & at & $40.8$ & $44.5$ & $41.4$ & $42.7$ & $46.3$ & $55.6$ & $41.8$ & $41.9$ & $53.7$ & $60.8$ & $45.0$ & $41.5$ & $44.8$ & $30.8$ & $31.9$ & $\mathbf{44.3}$ \\
    Gong~\textit{et al.} '21~\cite{gongPoseAugDifferentiablePose2021} & ag & - & - & - & - & - & - & - & - & - & - & - & - & - & - & - & $50.2$ \\
    \midrule
    ours $C$+$H$+$M$ & ag & $49.2$ & $51.9$ & $49.6$ & $50.9$ & $55.4$ & $60.4$ & $45.4$ & $48.8$ & $64.3$ & $75.2$ & $53.0$ & $47.3$ & $54.0$ & $39.2$ & $45.5$ & $\mathbf{52.7}$ \\
    ours $C$+$H$+$S$+$M$ & ag & $51.2$ & $51.9$ & $49.7$ & $52.1$ & $56.0$ & $60.3$ & $47.1$ & $48.8$ & $62.8$ & $75.6$ & $53.1$ & $48.2$ & $54.2$ & $40.5$ & $47.3$ & $53.3$ \\
    \midrule
    ours $C$+$H$+$M$ & g & $50.4$ & $53.8$ & $50.8$ & $52.9$ & $57.5$ & $62.9$ & $47.2$ & $50.5$ & $65.8$ & $78.6$ & $54.8$ & $49.2$ & $56.0$ & $40.7$ & $46.3$ & $54.5$ \\
    ours $C$+$H$+$S$+$M$ & g & $52.4$ & $53.9$ & $50.8$ & $54.0$ & $57.8$ & $62.6$ & $48.6$ & $50.3$ & $64.6$ & $79.0$ & $54.7$ & $49.6$ & $56.3$ & $41.6$ & $48.1$ & $54.9$ \\
  \end{tabular}
  }
  \caption[H36M Results]{Results on the H36M dataset reported as mean per joint position error (MPJPE). Legend of properties (Prop.), influencing the results: (a) used test-time augmentation (in ours only flip test is applied), (g) used ground truth bounding boxes, (s) sampled every 64th frame of validation set, (t) used temporal information}
  \label{tab:h36m_eval_results}
\end{table*}

For 3D human pose estimation, we first consider the performance compared to other methods on the H36M validation dataset. The results of 3D pose estimation are shown in Table~\ref{tab:h36m_eval_results}. It summarizes further approaches, which differ in the methods used, input data, and test methods. Some approaches use test-time augmentations like flip-testing. Others use temporal information to improve the results. Our novel PedRecNet is based on single-frame estimation. For better comparability, we also show results using flip-testing in addition to the results without any form of test-time augmentation. The performance of our method with an average MPJPE of $52.7mm$ is comparable to current SOTA approaches such as Gong~\textit{et al.}~\cite{gongPoseAugDifferentiablePose2021} with $50.2mm$ and Luvizon~\textit{et al.}~\cite{luvizonMultitaskDeepLearning2020} with $48.6mm$. The use of temporal information leads to noticeable better results in this benchmark. This becomes clear when comparing the two results of Pavllo~\textit{et al.}~\cite{pavllo3DHumanPose2019}. An improvement of $4.7mm$ could be achieved by using temporal information. The performance from PedRecNet, depending on the datasets used, is very similar but decreases slightly when simulation data is added. However, this may also be because the simulation data is partly very different from the H36M data in terms of distances and body size of the persons.

\begin{figure*}[htp]
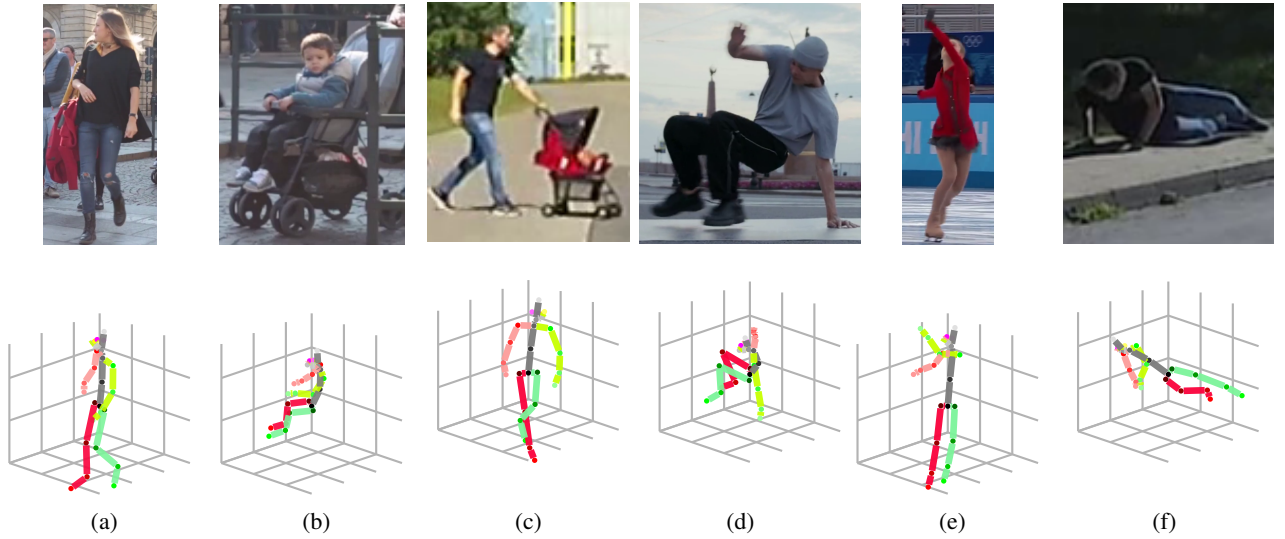

  \centering
  \subcaptionbox{\label{fig:example_3d_skeletons:a}}[.15\textwidth]{%
    \input{gfx/pedrec/example_3d_skeleton_1.tex}
  }
  \subcaptionbox{\label{fig:example_3d_skeletons:b}}[.15\textwidth]{%
    \input{gfx/pedrec/example_3d_skeleton_2.tex}
  }
  \subcaptionbox{\label{fig:example_3d_skeletons:c}}[.15\textwidth]{%
    \input{gfx/pedrec/example_3d_skeleton_4.tex}
  }
  \subcaptionbox{\label{fig:example_3d_skeletons:d}}[.15\textwidth]{%
  \input{gfx/pedrec/example_3d_skeleton_3.tex}
}
\subcaptionbox{\label{fig:example_3d_skeletons:e}}[.15\textwidth]{%
\input{gfx/pedrec/example_3d_skeleton_5.tex}
}
\subcaptionbox{\label{fig:example_3d_skeletons:f}}[.15\textwidth]{%
\input{gfx/pedrec/example_3d_skeleton_6.tex}
}
  \caption[3D Human pose estimation in the wild: PedRecNet examples]{3D Human pose estimation \enquote{in the wild}: PedRecNet examples. Top: Cropped image of the person inputed in the network. Bottom: Predicted 3D human pose.}
  \label{fig:example_3d_skeletons}
\end{figure*}

\begin{figure}[htp]
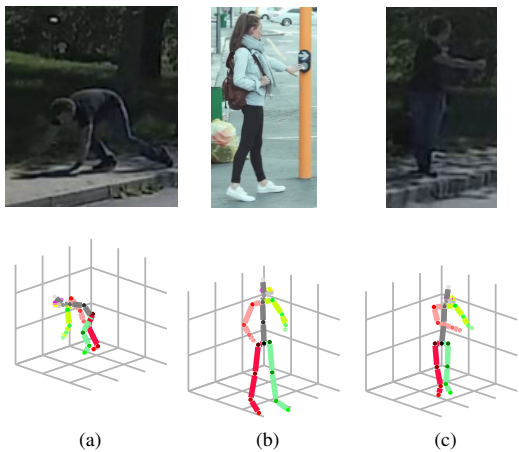

  \centering
  \resizebox*{0.8\columnwidth}{!}{%
  \subcaptionbox{\label{fig:example_3d_skeleton_false_dets:b}}[.15\textwidth]{%
    \input{gfx/pedrec/example_3d_skeleton_wrong_2.tex}
  }
\subcaptionbox{\label{fig:example_3d_skeleton_false_dets:e}}[.15\textwidth]{%
\input{gfx/pedrec/example_3d_skeleton_wrong_6.tex}
}
\subcaptionbox{\label{fig:example_3d_skeleton_false_dets:f}}[.15\textwidth]{%
\input{gfx/pedrec/example_3d_skeleton_wrong_3.tex}
}}
\caption[3D Human pose estimation in the wild: PedRecNet false predictions]{3D Human pose estimation in the wild: PedRecNet false predictions. Top: Cropped image of the person inputed in the network. Bottom: Predicted 3D human pose.}
  \label{fig:example_3d_skeleton_false_dets}
\end{figure}

Figure~\ref{fig:example_3d_skeletons} shows some examples on \enquote{in the wild} real data. The examples are from various sources and include different cameras, focal lengths, exposures and perspectives. Examples \ref{fig:example_3d_skeletons:d}-\ref{fig:example_3d_skeletons:f} show that even in challenging situations a good 3D pose can be predicted. In contrast, we have reported some error cases in Figure \ref{fig:example_3d_skeleton_false_dets}. Figure \ref{fig:example_3d_skeleton_false_dets:b} shows an extreme corner case where the pose detection fails completely. Note that the corner case dataset from our work~\cite{ludlUsingSimulationImprove2018} was not used during training. In Figure \ref{fig:example_3d_skeleton_false_dets:e} the pose is correct in principle, but the outstretched left arm is not correctly recognized. From the pose only, it is not clear that the person is just operating a traffic light switch. Example \ref{fig:example_3d_skeleton_false_dets:f} shows a false recognition due to self occlusion. The body occludes the left arm, but the 3D pose recognition estimates it to be hidden behind the right arm and displays it stretched out accordingly. These false detections by occlusion could possibly be improved by further training data or the use of temporal context.

\begin{figure}
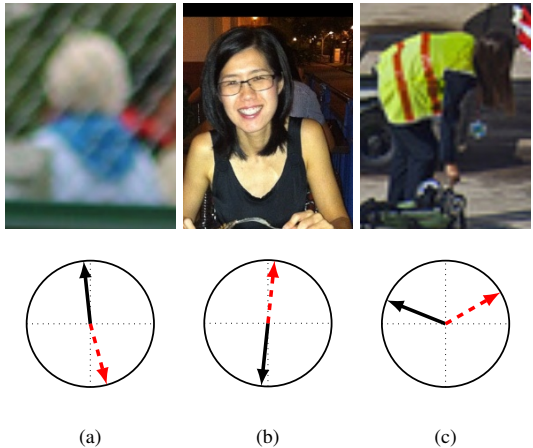

  \centering
  \resizebox*{0.8\columnwidth}{!}{%
  \subcaptionbox{\label{fig:mebow_wrong_labels:a}}[.15\textwidth]{%
    \input{gfx/pedrec/mebow_false_examples/wrong_label_1.tex}
  }
  \subcaptionbox{\label{fig:mebow_crowded_unclears:a}}[.15\textwidth]{%
  \input{gfx/pedrec/mebow_false_examples/crowded_unclear_1.tex}
  }
  \subcaptionbox{\label{fig:mebow_wrong_predicted:c}}[.15\textwidth]{%
  \input{gfx/pedrec/mebow_false_examples/wrong_predicted_3.tex}
  }}
  \caption[MEBOW validation dataset: Wrong labels]{MEBOW validation dataset: Examples of misclassifications. The red dotted arrow shows the annotated ground truth, the black arrow shows the prediction of PedRecNet. The misclassifications are caused by: (a) false ground truth label, (b) occlusion of the labeled person (the one in the back of the woman) and (c) PedRecNet misclassification.}
  \label{fig:mebow_wrong_labels}
\end{figure}

\paragraph{Body orientation estimation}
\label{sec:pedrec:results:orientation}
As described in the related work section, there are only few datasets in the area of body and head orientation estimation for full-body inputs. For a state-of-the-art comparison, we found the relatively new dataset ~\cite{wuMEBOWMonocularEstimation2020} suitable, which provides body orientation labels for the COCO dataset. However, it only contains the azimuthal angle $\varphi$ and labels for the head pose are not included. We also used the TUD~\cite{andrilukaMonocular3DPose2010} dataset in the analysis, although it only contains $309$ samples in the validation dataset. Accordingly, the significance of the results here is relatively low. Table~\ref{tab:pedrec_body_orientation_estimation_results} gives an overview on the results on these datasets. It is to notice that with the PedRecNet we already achieve an $Acc. (22.5^\circ)$ of $75.4\%$ on the TUD~\cite{andrilukaMonocular3DPose2010} dataset and $80.2\%$ on the MEBOW dataset. For $Acc. (45^\circ)$, which is often sufficient for real-world applications, we even achieve $98.1\%$ on the TUD~\cite{andrilukaMonocular3DPose2010} dataset and $94.7\%$ on the MEBOW dataset. These are surprising results for not using any training data from the corresponding training datasets. Especially when compared to the earlier approaches of Hara~\textit{et al.}~\cite{haraDesigningDeepConvolutional2017} and Yu~\textit{et al.}~\cite{yuContinuousPedestrianOrientation2019}, the PedRecNet gives better results without ever having seen any data from the TUD~\cite{andrilukaMonocular3DPose2010} dataset. We are accordingly able to provide a solid baseline here purely with simulated data. It should be noted, however, that 3D pose data is also used for orientation estimation and the training for this has included real data from the H36M dataset. When the MEBOW training data are used in addition to the simulation data, the $Acc. (22.5^\circ)$ and $Acc. (45^\circ)$ improve by $11.5\%$ and $2.3\%$, respectively, and are $2.2\%$ and $1.2\%$ worse than the results reported by Wu~\textit{et al.}. In total, $159$ body orientations were predicted with an error above $45^\circ$. We analyzed these misclassifications further and detected erroneous ground truth labels for $37$ images, some of which are shown in Figure~\ref{fig:mebow_wrong_labels}. 

\begin{table*}[!htbp]
  \centering

  \begin{tabular}{l l l l l l} \toprule
      Network & Trainset & Testset & $Acc. (22.5^\circ)$ & $Acc. (45^\circ)$ & MAE($^\circ$) \\ 
      \midrule
      Wu~\textit{et al.}\cite{wuMEBOWMonocularEstimation2020} (2020) & MEBOW & MEBOW & $\mathbf{93.9}$ & $\mathbf{98.2}$ & $8.4$ \\
      ours & MEBOW & MEBOW & $\mathbf{92.3}$ & $\mathbf{97.0}$ & $9.7$ \\
      ours & SIM+MEBOW & MEBOW & $91.7$ & $\mathbf{97.0}$ & $10.0$ \\
      ours & SIM & MEBOW & $80.2$ & $94.7$ & $16.1$ \\\midrule
      Hara~\textit{et al.}\cite{haraDesigningDeepConvolutional2017} (2017) & TUD & TUD & $70.6$ & $86.1$ & $26.6$ \\
      Yu~\textit{et al.}\cite{yuContinuousPedestrianOrientation2019} (2019) & TUD & TUD & $75.7$ & $96.8$ & $15.3$ \\
      Wu~\textit{et al.}\cite{wuMEBOWMonocularEstimation2020} (2020) & MEBOW & TUD & $77.3$ & $99.0$ & $14.3$ \\
      ours & MEBOW & TUD & $\mathbf{79.6}$ & $\mathbf{99.0}$ & $10.8$ \\
      ours & SIM+MEBOW & TUD & $\mathbf{77.3}$ & $\mathbf{98.7}$ & $14.3$ \\
      ours & SIM & TUD & $\mathbf{75.4}$ & $\mathbf{98.1}$ & $16.0$ \\
      \midrule
      ours & MEBOW & SIM-C01 & $76.2$ & $97.0$ & $16.6$ \\
      ours & SIM+MEBOW & SIM-C01 & $\mathbf{79.7}$ & $\mathbf{97.9}$ & $15.3$ \\
      ours & SIM & SIM-C01 & $78.7$ & $96.5$ & $16.0$ \\
  \end{tabular}
  \caption[Human body orientation ($\varphi$) results]{Human body orientation ($\varphi$) test results on the MEBOW, TUD~\cite{andrilukaMonocular3DPose2010} and SIM-C01V datasets. The column trainset specifies the training dataset(s) used to train the specific networks. Testset specifies on which testsets the results are reported on. In addition to the accuracy in $22.5^{\circ}$ and $45^{\circ}$ intervals we report the mean average error (MAE).}
  \label{tab:pedrec_body_orientation_estimation_results}
\end{table*}

\paragraph{Head orientation estimation}
For the head orientation ($\varphi$) estimation, we use the SIM-C01V dataset. We consider only the $\varphi$ estimate at this point because $\theta$ is underrepresented in the SIM-C01 dataset; the head orientations are relatively horizontal in the pedestrian actions in almost all cases. Accordingly, for the estimation of $\theta$, further targeted experiments and new data recordings are needed in the future. The results for the estimation of the head $\varphi$ orientation are shown in Table~\ref{tab:pedrec_head_orientation_estimation_results}.

\begin{table}[!htbp]
  \centering
  \resizebox{1\columnwidth}{!}{%
  \begin{tabular}{l l l l l l} \toprule
      Network & Trainset & Testset & $Acc. (22.5^\circ)$ & $Acc. (45^\circ)$ & $MAE$ \\ \midrule
      PedRec & SIM+MEBOW & SIM-C01V & $77.1$ & $95.1$ & $16.65$ \\
      PedRec & SIM & SIM-C01V & $76.3$ & $94.8$ & $17.43$ \\
  \end{tabular}}
  \caption[PedRecNet: Head orientation results]{PedRecNet: Head orientation test results for $\varphi$.}
  \label{tab:pedrec_head_orientation_estimation_results}
\end{table}

\begin{figure}
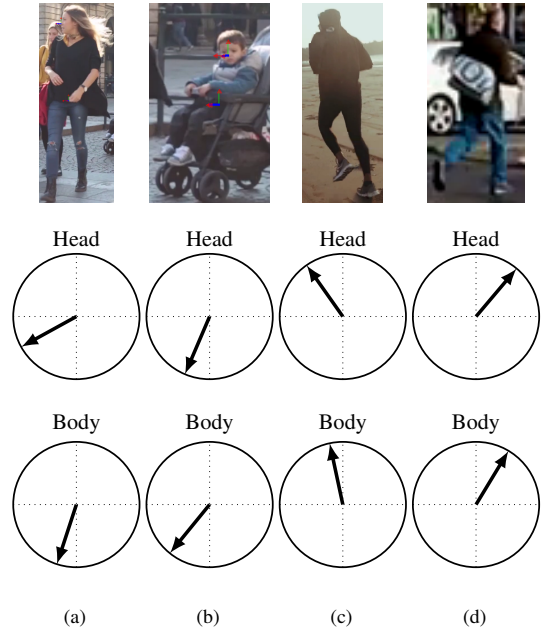

  \centering
  \resizebox*{0.8\columnwidth}{!}{%
  \subcaptionbox{\label{fig:orientation_inthewild_examples:a}}[.23\columnwidth]{%
  \input{gfx/pedrec/orientation_example_1.tex}
}
  \subcaptionbox{\label{fig:orientation_inthewild_examples:b}}[.23\columnwidth]{%
    \input{gfx/pedrec/orientation_example_2.tex}
  }
  \subcaptionbox{\label{fig:orientation_inthewild_examples:c}}[.23\columnwidth]{%
    \input{gfx/pedrec/orientation_example_3.tex}
  }
  \subcaptionbox{\label{fig:orientation_inthewild_examples:d}}[.23\columnwidth]{%
    \input{gfx/pedrec/orientation_example_4.tex}
  }}
  \caption[Head and body orientation estimation \enquote{in the wild}: Examples]{Head and body orientation estimation \enquote{in the wild}: Examples. Top: Cropped image of the person processed by the network. Bottom two rows: Predicted head and body orientation $\varphi$.}
  \label{fig:orientation_inthewild_examples}
\end{figure}

The results are slightly inferior to the body orientation estimation by $2.6\%$ and $2.8\%$ for $Acc. (22.5^\circ)$ and $Acc. (45^\circ)$, respectively. In general, however, performance on the body and head orientation estimates is relatively similar. The somewhat inferior performance can be explained by the head region's smaller image area than the body region. 
We are not currently aware of a larger and publicly dataset that includes head orientation images in addition to full-body images. Therefore, most approaches to head orientation estimation work with datasets that only contain cropped faces. In productive applications, face recognition can then be performed first, followed by a crop of the face, and orientation estimation can be performed based on this cropped face bounding box. In addition, most datasets only contain faces, which means that a side view or the back of the head cannot usually be used for orientation estimation. In our approach, the entire body is always considered, which enables head orientation estimation even for a side and back view of a person. However, based on subjective observation of \enquote{in-the-wild} examples, we think that we can achieve similar performance on real data for head pose recognition as for body pose recognition when trained on simulation data only. We show \enquote{in-the-wild} examples in Figure~\ref{fig:orientation_inthewild_examples}.

Figure~\ref{fig:orientation_inthewild_examples:a} shows a typical example, where one can nicely depict the different estimates of head versus body orientation. Example~\ref{fig:orientation_inthewild_examples:b} shows a boy in a stroller, which shows that the orientation estimation gives good results even in non-upright positions. Figure \ref{fig:orientation_inthewild_examples:c} shows a person who was photographed from behind. Especially the correct head pose estimation is interesting, although the person wears a hood and only a small part of the nose is visible. Another interesting example is demonstrated in Figure \ref{fig:orientation_inthewild_examples:d}, where the orientation estimation is based on input data of a person shot from behind and only visible in a low-resolution image section of about $38\times78px$. 

\section{Conclusion}
With PedRecNet, we presented a simple yet efficient architecture that performs multiple tasks simultaneously and can run on consumer hardware at over 15FPS even with multiple people. The network achieves performance that is comparable to current SOTA methods for 2D and 3D pose detection and orientation estimation. Our model combines all these tasks in a simple and extensible architecture which is straight forward to train. Thus, the introduced model is also well suited as a baseline for further research. We have further shown that we can train the orientation estimation purely with simulation data and achieve high accuracy on real data without requiring real sensor data for training.

\section*{ACKNOWLEDGMENT}
This project has been supported by the Continental AG as part of a research cooperation.


\bibliography{literature.bib}

\begin{thebibliography}{10}
\providecommand{\url}[1]{#1}
\csname url@rmstyle\endcsname
\providecommand{\newblock}{\relax}
\providecommand{\bibinfo}[2]{#2}
\providecommand\BIBentrySTDinterwordspacing{\spaceskip=0pt\relax}
\providecommand\BIBentryALTinterwordstretchfactor{4}
\providecommand\BIBentryALTinterwordspacing{\spaceskip=\fontdimen2\font plus
\BIBentryALTinterwordstretchfactor\fontdimen3\font minus
  \fontdimen4\font\relax}
\providecommand\BIBforeignlanguage[2]{{%
\expandafter\ifx\csname l@#1\endcsname\relax
\typeout{** WARNING: IEEEtran.bst: No hyphenation pattern has been}%
\typeout{** loaded for the language `#1'. Using the pattern for}%
\typeout{** the default language instead.}%
\else
\language=\csname l@#1\endcsname
\fi
#2}}

\bibitem{ludlSimpleEfficientRealtime2019}
D.~Ludl, T.~Gulde, and C.~Curio, ``Simple yet efficient real-time pose-based
  action recognition,'' in \emph{22nd {{IEEE Int}}. {{Conf}}. on {{Intelligent
  Transportation Systems}} ({{ITSC}})}, Nov. 2019, pp. 581--588.

\bibitem{bulthoffTopdownInfluencesStereoscopic1998}
I.~B{\"u}lthoff, H.~B{\"u}lthoff, and P.~Sinha, ``Top-down influences on
  stereoscopic depth-perception,'' \emph{Nature Neuroscience}, vol.~1, no.~3,
  pp. 254--257, July 1998.

\bibitem{ludlEnhancingDataDrivenAlgorithms2020}
D.~Ludl, T.~Gulde, and C.~Curio, ``Enhancing {{Data-Driven Algorithms}} for
  {{Human Pose Estimation}} and {{Action Recognition Through Simulation}},''
  \emph{IEEE Transactions on Intelligent Transportation Systems}, pp. 1--10,
  2020.

\bibitem{mehtaVNectRealtime3D2017}
D.~Mehta, S.~Sridhar, O.~Sotnychenko, H.~Rhodin, M.~Shafiei, H.-P. Seidel,
  W.~Xu, D.~Casas, and C.~Theobalt, ``{{VNect}}: Real-time {{3D}} human pose
  estimation with a single {{RGB}} camera,'' \emph{ACM Transactions on
  Graphics}, vol.~36, no.~4, pp. 1--14, July 2017.

\bibitem{mehtaSingleShotMultiperson3D2018}
D.~Mehta, O.~Sotnychenko, F.~Mueller, W.~Xu, S.~Sridhar, G.~{Pons-Moll}, and
  C.~Theobalt, ``Single-{{Shot Multi-person 3D Pose Estimation}} from
  {{Monocular RGB}},'' in \emph{2018 {{International Conference}} on {{3D
  Vision}} ({{3DV}})}.\hskip 1em plus 0.5em minus 0.4em\relax {Verona}: {IEEE},
  Sept. 2018, pp. 120--130.

\bibitem{mehtaXNectRealtimeMultiperson2020}
D.~Mehta, O.~Sotnychenko, F.~Mueller, W.~Xu, M.~Elgharib, P.~Fua, H.-P. Seidel,
  H.~Rhodin, G.~{Pons-Moll}, and C.~Theobalt, ``{{XNect}}: Real-time
  multi-person {{3D}} motion capture with a single {{RGB}} camera,'' \emph{ACM
  Transactions on Graphics}, vol.~39, no.~4, July 2020.

\bibitem{luvizonMultitaskDeepLearning2020}
D.~C. Luvizon, H.~Tabia, and D.~Picard, ``Multi-task {{Deep Learning}} for
  {{Real-Time 3D Human Pose Estimation}} and {{Action Recognition}},''
  \emph{IEEE Transactions on Pattern Analysis and Machine Intelligence}, pp.
  1--1, 2020.

\bibitem{chen3DHumanPose2017}
C.-H. Chen and D.~Ramanan, ``{{3D Human Pose Estimation}} = {{2D Pose
  Estimation}} + {{Matching}},'' in \emph{2017 {{IEEE Conference}} on
  {{Computer Vision}} and {{Pattern Recognition}} ({{CVPR}})}.\hskip 1em plus
  0.5em minus 0.4em\relax {Honolulu, HI}: {IEEE}, July 2017, pp. 5759--5767.

\bibitem{pavllo3DHumanPose2019}
D.~Pavllo, C.~Feichtenhofer, D.~Grangier, and M.~Auli, ``{{3D}} human pose
  estimation in video with temporal convolutions and semi-supervised
  training,'' in \emph{Conference on {{Computer Vision}} and {{Pattern
  Recognition}} ({{CVPR}})}, 2019.

\bibitem{tekinLearningFuse2D2017}
B.~Tekin, P.~{Marquez-Neila}, M.~Salzmann, and P.~Fua, ``Learning to {{Fuse
  2D}} and {{3D Image Cues}} for {{Monocular Body Pose Estimation}},'' in
  \emph{2017 {{IEEE International Conference}} on {{Computer Vision}}
  ({{ICCV}})}.\hskip 1em plus 0.5em minus 0.4em\relax {Venice}: {IEEE}, Oct.
  2017, pp. 3961--3970.

\bibitem{zhou3DHumanPose2017}
X.~Zhou, Q.~Huang, X.~Sun, X.~Xue, and Y.~Wei, ``Towards {{3D Human Pose
  Estimation}} in the {{Wild}}: {{A Weakly-Supervised Approach}},'' in
  \emph{2017 {{IEEE International Conference}} on {{Computer Vision}}
  ({{ICCV}})}.\hskip 1em plus 0.5em minus 0.4em\relax {Venice}: {IEEE}, Oct.
  2017, pp. 398--407.

\bibitem{kolotourosLearningReconstruct3D2019}
N.~Kolotouros, G.~Pavlakos, M.~Black, and K.~Daniilidis, ``Learning to
  {{Reconstruct 3D Human Pose}} and {{Shape}} via {{Model-Fitting}} in the
  {{Loop}},'' in \emph{2019 {{IEEE}}/{{CVF International Conference}} on
  {{Computer Vision}} ({{ICCV}})}.\hskip 1em plus 0.5em minus 0.4em\relax
  {Seoul, Korea (South)}: {IEEE}, Oct. 2019, pp. 2252--2261.

\bibitem{bogoKeepItSMPL2016}
F.~Bogo, A.~Kanazawa, C.~Lassner, P.~Gehler, J.~Romero, and M.~J. Black, ``Keep
  {{It SMPL}}: {{Automatic Estimation}} of {{3D Human Pose}} and {{Shape}} from
  a {{Single Image}},'' in \emph{Computer {{Vision}} \textendash{} {{ECCV}}
  2016}, B.~Leibe, J.~Matas, N.~Sebe, and M.~Welling, Eds.\hskip 1em plus 0.5em
  minus 0.4em\relax {Cham}: {Springer International Publishing}, 2016, vol.
  9909, pp. 561--578.

\bibitem{zhouDeepKinematicPose2016}
X.~Zhou, X.~Sun, W.~Zhang, S.~Liang, and Y.~Wei, ``Deep {{Kinematic Pose
  Regression}},'' in \emph{Computer {{Vision}} \textendash{} {{ECCV}} 2016
  {{Workshops}}}, G.~Hua and H.~J{\'e}gou, Eds.\hskip 1em plus 0.5em minus
  0.4em\relax {Cham}: {Springer International Publishing}, 2016, vol. 9915, pp.
  186--201.

\bibitem{luvizon2D3DPose2018}
D.~C. Luvizon, D.~Picard, and H.~Tabia, ``{{2D}}/{{3D Pose Estimation}} and
  {{Action Recognition Using Multitask Deep Learning}},'' in \emph{2018
  {{IEEE}}/{{CVF Conference}} on {{Computer Vision}} and {{Pattern
  Recognition}}}.\hskip 1em plus 0.5em minus 0.4em\relax {Salt Lake City, UT,
  USA}: {IEEE}, June 2018, pp. 5137--5146.

\bibitem{martinezSimpleEffectiveBaseline2017}
J.~Martinez, R.~Hossain, J.~Romero, and J.~J. Little, ``A {{Simple Yet
  Effective Baseline}} for 3d {{Human Pose Estimation}},'' in \emph{2017 {{IEEE
  International Conference}} on {{Computer Vision}} ({{ICCV}})}.\hskip 1em plus
  0.5em minus 0.4em\relax {Venice}: {IEEE}, Oct. 2017, pp. 2659--2668.

\bibitem{loperSMPLSkinnedMultiperson2015}
M.~Loper, N.~Mahmood, J.~Romero, G.~{Pons-Moll}, and M.~J. Black, ``{{SMPL}}:
  {{A Skinned Multi-person Linear Model}},'' \emph{ACM Trans. Graph.}, vol.~34,
  no.~6, pp. 1--16, Oct. 2015.

\bibitem{andrilukaPictorialStructuresRevisited2009}
M.~Andriluka, S.~Roth, and B.~Schiele, ``Pictorial structures revisited:
  {{People}} detection and articulated pose estimation,'' in \emph{2009 {{IEEE
  Conference}} on {{Computer Vision}} and {{Pattern Recognition}}}, June 2009,
  pp. 1014--1021.

\bibitem{enzweilerIntegratedPedestrianClassification2010}
M.~Enzweiler and D.~M. Gavrila, ``Integrated pedestrian classification and
  orientation estimation,'' in \emph{2010 {{IEEE Computer Society Conference}}
  on {{Computer Vision}} and {{Pattern Recognition}}}.\hskip 1em plus 0.5em
  minus 0.4em\relax {San Francisco, CA, USA}: {IEEE}, June 2010, pp. 982--989.

\bibitem{flohrJointProbabilisticPedestrian2014}
F.~Flohr, M.~{Dumitru-Guzu}, J.~F.~P. Kooij, and D.~M. Gavrila, ``Joint
  probabilistic pedestrian head and body orientation estimation,'' in
  \emph{2014 {{IEEE Intelligent Vehicles Symposium Proceedings}}}.\hskip 1em
  plus 0.5em minus 0.4em\relax {MI, USA}: {IEEE}, June 2014, pp. 617--622.

\bibitem{flohrProbabilisticFrameworkJoint2015}
------, ``A {{Probabilistic Framework}} for {{Joint Pedestrian Head}} and
  {{Body Orientation Estimation}},'' \emph{IEEE Transactions on Intelligent
  Transportation Systems}, vol.~16, no.~4, pp. 1872--1882, Aug. 2015.

\bibitem{guptaNoseEyesEars2019}
A.~Gupta, K.~Thakkar, V.~Gandhi, and P.~J. Narayanan, ``Nose, {{Eyes}} and
  {{Ears}}: {{Head Pose Estimation}} by {{Locating Facial Keypoints}},'' in
  \emph{{{ICASSP}} 2019 - 2019 {{IEEE International Conference}} on
  {{Acoustics}}, {{Speech}} and {{Signal Processing}} ({{ICASSP}})}.\hskip 1em
  plus 0.5em minus 0.4em\relax {Brighton, United Kingdom}: {IEEE}, May 2019,
  pp. 1977--1981.

\bibitem{ruizFineGrainedHeadPose2018}
N.~Ruiz, E.~Chong, and J.~M. Rehg, ``Fine-{{Grained Head Pose Estimation
  Without Keypoints}},'' in \emph{2018 {{IEEE}}/{{CVF Conference}} on
  {{Computer Vision}} and {{Pattern Recognition Workshops}} ({{CVPRW}})}.\hskip
  1em plus 0.5em minus 0.4em\relax {Salt Lake City, UT, USA}: {IEEE}, June
  2018, pp. 2155--215\,509.

\bibitem{panSelfPacedDeepRegression2020}
L.~Pan, S.~Ai, Y.~Ren, and Z.~Xu, ``Self-{{Paced Deep Regression Forests}} with
  {{Consideration}} on {{Underrepresented Examples}},'' in \emph{Computer
  {{Vision}} \textendash{} {{ECCV}} 2020}, A.~Vedaldi, H.~Bischof, T.~Brox, and
  J.-M. Frahm, Eds.\hskip 1em plus 0.5em minus 0.4em\relax {Cham}: {Springer
  International Publishing}, 2020, vol. 12375, pp. 271--287.

\bibitem{huDeepConvolutionalNeural2021}
Z.~Hu, Y.~Xing, C.~Lv, P.~Hang, and J.~Liu, ``Deep convolutional neural
  network-based {{Bernoulli}} heatmap for head pose estimation,''
  \emph{Neurocomputing}, vol. 436, pp. 198--209, May 2021.

\bibitem{valleMultitaskHeadPose2020}
R.~Valle, J.~M. Buenaposada, and L.~Baumela, ``Multi-task head pose estimation
  in-the-wild,'' \emph{IEEE Transactions on Pattern Analysis and Machine
  Intelligence}, pp. 1--1, 2020.

\bibitem{xiaEfficientMultitaskNeural2021}
J.~Xia, H.~Zhang, S.~Wen, S.~Yang, and M.~Xu, ``An {{Efficient Multitask Neural
  Network}} for {{Face Alignment}}, {{Head Pose Estimation}} and {{Face
  Tracking}},'' \emph{arXiv:2103.07615 [cs]}, Mar. 2021.

\bibitem{heoEstimationPedestrianPose2019}
D.~Heo, J.~Nam, and B.~Ko, ``Estimation of {{Pedestrian Pose Orientation Using
  Soft Target Training Based}} on {{Teacher}}\textendash{{Student
  Framework}},'' \emph{Sensors}, vol.~19, no.~5, p. 1147, Mar. 2019.

\bibitem{leeHeadBodyOrientation2019}
D.~Lee, M.-H. Yang, and S.~Oh, ``Head and {{Body Orientation Estimation Using
  Convolutional Random Projection Forests}},'' \emph{IEEE Transactions on
  Pattern Analysis and Machine Intelligence}, vol.~41, no.~1, pp. 107--120,
  Jan. 2019.

\bibitem{steinhoffPedestrianHeadBody2020}
M.~Steinhoff and D.~G{\"o}hring, ``Pedestrian {{Head}} and {{Body Pose
  Estimation}} with {{CNN}} in the {{Context}} of {{Automated Driving}}:,'' in
  \emph{Proceedings of the 6th {{International Conference}} on {{Vehicle
  Technology}} and {{Intelligent Transport Systems}}}.\hskip 1em plus 0.5em
  minus 0.4em\relax {Prague, Czech Republic}: {SCITEPRESS - Science and
  Technology Publications}, 2020, pp. 353--360.

\bibitem{wuMEBOWMonocularEstimation2020}
C.~Wu, Y.~Chen, J.~Luo, C.-C. Su, A.~Dawane, B.~Hanzra, Z.~Deng, B.~Liu, J.~Z.
  Wang, and C.-h. Kuo, ``{{MEBOW}}: {{Monocular Estimation}} of {{Body
  Orientation}} in the {{Wild}},'' in \emph{2020 {{IEEE}}/{{CVF Conference}} on
  {{Computer Vision}} and {{Pattern Recognition}} ({{CVPR}})}.\hskip 1em plus
  0.5em minus 0.4em\relax {Seattle, WA, USA}: {IEEE}, June 2020, pp.
  3448--3458.

\bibitem{xiaoSimpleBaselinesHuman2018}
B.~Xiao, H.~Wu, and Y.~Wei, ``Simple {{Baselines}} for {{Human Pose
  Estimation}} and {{Tracking}},'' in \emph{European {{Conference}} on
  {{Computer Vision}} ({{ECCV}})}.\hskip 1em plus 0.5em minus 0.4em\relax
  {Springer International Publishing}, 2018, pp. 472--487.

\bibitem{internationalorganizationforstandardizationISO80000220192019}
{International Organization for Standardization}, ``{{ISO}} 80000-2:2019 -
  {{Quantities}} and units \textemdash{} {{Part}} 2: {{Mathematics}},'' Tech.
  Rep., Aug. 2019.

\bibitem{linMicrosoftCOCOCommon2014}
T.-Y. Lin, M.~Maire, S.~Belongie, J.~Hays, P.~Perona, D.~Ramanan,
  P.~Doll{\'a}r, and C.~L. Zitnick, ``Microsoft {{COCO}}: {{Common Objects}} in
  {{Context}},'' in \emph{European {{Conference}} on {{Computer Vision}}
  ({{ECCV}})}.\hskip 1em plus 0.5em minus 0.4em\relax {Springer International
  Publishing}, 2014, pp. 740--755.

\bibitem{ionescuHuman36MLarge2014}
C.~Ionescu, P.~Papava, V.~Olaru, and C.~Sminchisescu, ``Human3.{{6M}}: {{Large
  Scale Datasets}} and {{Predictive Methods}} for {{3D Human Sensing}} in
  {{Natural Environments}},'' \emph{IEEE Transactions on Pattern Analysis and
  Machine Intelligence}, vol.~36, no.~7, pp. 1325--1339, July 2014.

\bibitem{andrilukaMonocular3DPose2010}
M.~Andriluka, S.~Roth, and B.~Schiele, ``Monocular {{3D}} pose estimation and
  tracking by detection,'' in \emph{2010 {{IEEE Computer Society Conference}}
  on {{Computer Vision}} and {{Pattern Recognition}}}.\hskip 1em plus 0.5em
  minus 0.4em\relax {San Francisco, CA, USA}: {IEEE}, June 2010, pp. 623--630.

\bibitem{kendallMultitaskLearningUsing2018}
A.~Kendall, Y.~Gal, and R.~Cipolla, ``Multi-task {{Learning Using Uncertainty}}
  to {{Weigh Losses}} for {{Scene Geometry}} and {{Semantics}},'' in \emph{2018
  {{IEEE}}/{{CVF Conference}} on {{Computer Vision}} and {{Pattern
  Recognition}}}.\hskip 1em plus 0.5em minus 0.4em\relax {Salt Lake City, UT,
  USA}: {IEEE}, June 2018, pp. 7482--7491.

\bibitem{loshchilovDecoupledWeightDecay2019}
I.~Loshchilov and F.~Hutter, ``Decoupled {{Weight Decay Regularization}},'' in
  \emph{7th {{International Conference}} on {{Learning Representations}}
  ({{ICLR}})}, Jan. 2019.

\bibitem{kingmaAdamMethodStochastic2015}
D.~P. Kingma and J.~Ba, ``Adam: {{A Method}} for {{Stochastic Optimization}},''
  \emph{arXiv:1412.6980 [cs]}, 2015.

\bibitem{smithCyclicalLearningRates2017a}
L.~N. Smith, ``Cyclical {{Learning Rates}} for {{Training Neural Networks}},''
  in \emph{2017 {{IEEE Winter Conference}} on {{Applications}} of {{Computer
  Vision}} ({{WACV}})}.\hskip 1em plus 0.5em minus 0.4em\relax {Santa Rosa, CA,
  USA}: {IEEE}, Mar. 2017, pp. 464--472.

\bibitem{smithDisciplinedApproachNeural2018}
------, ``A disciplined approach to neural network hyper-parameters: {{Part}} 1
  -- learning rate, batch size, momentum, and weight decay,''
  \emph{arXiv:1803.09820 [cs, stat]}, Apr. 2018.

\bibitem{luvizonHumanPoseRegression2017}
D.~C. Luvizon, H.~Tabia, and D.~Picard, ``Human {{Pose Regression}} by
  {{Combining Indirect Part Detection}} and {{Contextual Information}},''
  \emph{Computers and Graphics}, vol.~85, pp. 15--22, Oct. 2017.

\bibitem{yang3DHumanPose2018a}
W.~Yang, W.~Ouyang, X.~Wang, J.~Ren, H.~Li, and X.~Wang, ``{{3D Human Pose
  Estimation}} in the {{Wild}} by {{Adversarial Learning}},'' in \emph{2018
  {{IEEE}}/{{CVF Conference}} on {{Computer Vision}} and {{Pattern
  Recognition}}}.\hskip 1em plus 0.5em minus 0.4em\relax {Salt Lake City, UT,
  USA}: {IEEE}, June 2018, pp. 5255--5264.

\bibitem{shanImprovingRobustnessAccuracy2021}
W.~Shan, H.~Lu, S.~Wang, X.~Zhang, and W.~Gao, ``Improving {{Robustness}} and
  {{Accuracy}} via {{Relative Information Encoding}} in {{3D Human Pose
  Estimation}},'' in \emph{Proceedings of the 29th {{ACM International
  Conference}} on {{Multimedia}}}.\hskip 1em plus 0.5em minus 0.4em\relax
  {Virtual Event China}: {ACM}, Oct. 2021, pp. 3446--3454.

\bibitem{gongPoseAugDifferentiablePose2021}
K.~Gong, J.~Zhang, and J.~Feng, ``{{PoseAug}}: {{A Differentiable Pose
  Augmentation Framework}} for {{3D Human Pose Estimation}},'' in
  \emph{Proceedings of the {{IEEE Conference}} on {{Computer Vision}} and
  {{Pattern Recognition}}}, {Virtual}, 2021, p.~10.

\bibitem{ludlUsingSimulationImprove2018}
D.~Ludl, T.~Gulde, S.~Thalji, and C.~Curio, ``Using {{Simulation}} to {{Improve
  Human Pose Estimation}} for {{Corner Cases}},'' in \emph{21st {{IEEE Int}}.
  {{Conf}}. on {{Intelligent Transportation Systems}} ({{ITSC}})}, Nov. 2018,
  pp. 3575--3582.

\bibitem{haraDesigningDeepConvolutional2017}
K.~Hara, R.~Vemulapalli, and R.~Chellappa, ``Designing {{Deep Convolutional
  Neural Networks}} for {{Continuous Object Orientation Estimation}},''
  \emph{arXiv:1702.01499 [cs]}, Feb. 2017.

\bibitem{yuContinuousPedestrianOrientation2019}
D.~Yu, H.~Xiong, Q.~Xu, J.~Wang, and K.~Li, ``Continuous {{Pedestrian
  Orientation Estimation}} using {{Human Keypoints}},'' in \emph{2019 {{IEEE
  International Symposium}} on {{Circuits}} and {{Systems}} ({{ISCAS}})}, May
  2019, pp. 1--5.

\end{thebibliography}
\bibliographystyle{IEEEtran}

\end{document}